\documentclass{sig-alternate-2013}

\usepackage{graphicx}
\usepackage{epstopdf}
\usepackage{color}
\usepackage{hyperref}
\usepackage[all]{hypcap}
\usepackage{url}
\usepackage{epstopdf}
\usepackage{listings}
\usepackage[square,numbers]{natbib}
\usepackage{tabu}
\usepackage[table]{xcolor}
\usepackage{enumitem}
\usepackage{hyphenat}
\newfont{\mycrnotice}{ptmr8t at 7pt}
 \newfont{\myconfname}{ptmri8t at 7pt}
 \let\crnotice\mycrnotice%
 \let\confname\myconfname%

\permission{Permission to make digital or hard copies of part or all of this work for personal or classroom use is granted without fee provided that copies are not made or distributed for profit or commercial advantage, and that copies bear this notice and the full citation on the first page. Copyrights for third-party components of this work must be honored. For all other uses, contact the owner/author(s). Copyright is held by the author/owner(s).}
 \conferenceinfo{WebSci'14,}{June 23--26, 2014, Bloomington, IN, USA.} 
 \copyrightetc{ACM \the\acmcopyr}
 \crdata{978-1-4503-2622-3/14/06. \\
 http://dx.doi.org/10.1145/2615569.2615639 }
 
\begin{document}

\title{Named Entity Evolution Analysis on Wikipedia}

\numberofauthors{2}

\author{
\alignauthor
Helge Holzmann\\
       \affaddr{L3S Research Center}\\
       \affaddr{Appelstr. 9a}\\
       \affaddr{30167 Hanover, Germany}\\
       \email{holzmann@L3S.de}
\alignauthor
Thomas Risse\\
       \affaddr{L3S Research Center}\\
       \affaddr{Appelstr. 9a}\\
       \affaddr{30167 Hanover, Germany}\\
       \email{risse@L3S.de}
}

\maketitle
\begin{abstract}
Accessing Web archives raises a number of issues caused by their temporal characteristics. Additional knowledge is needed to find and understand older texts. Especially entities mentioned in texts are subject to change. Most severe in terms of information retrieval are name changes. In order to find entities that have changed their name over time, search engines need to be aware of this evolution. We tackle this problem by analyzing Wikipedia in terms of entity evolutions mentioned in articles. We present statistical data on excerpts covering name changes, which will be used to discover similar text passages and extract evolution knowledge in future work.
\end{abstract}





\section{Introduction}

\begin{sloppypar}
With the Web as our daily source for updates and news, archiving of Web content has become important for preserving knowledge. However, due to the vast amount of data published at different times, accessing Web archives raises a number of challenges. Historical and evolution knowledge is essential to understand archived texts that were created a longer time ago. This is particularly important for finding entities as several characteristics may have changed over time. Most severe in this regard are changes of names, which are typically used as queries.
\end{sloppypar}

\citet{Berberich2009} tackled this problem by proposing a query reformulation technique to translate terms used in a query into terms used in older texts by connecting terms through their co-occurrence context today and in the past. \citet{Kaluarachchi2010} proposed another approach for computing temporally and semantically related terms using machine learning techniques on verbs shared among them. Both approaches work on the available corpus without exploiting explicit evolution information. \citet{Tahmasebi2012} proposed a similar approach that additionally considers the time of a name change. External resources have widely been neglected in this field. Even though several knowledge bases exist, most of them represent just current facts of entities. None of the popular ones, like DBpedia\footnote{\url{http://www.DBpedia.org}}, provide evolution information. An often used resource is semi-structured information from Wikipedia, such as info boxes. While these sometimes include alternative names of entities, they do not provide further details, for instance, whether or not a name is still valid and when a name was introduced. The only evolution that has been explicitly made available on Wikipedia is its revision history. However, this shows the development of an article rather than of the corresponding entity. \citet{Kanhabua2010} achieved promising results in discovering former names from anchor texts by exploiting the actual revisions of articles. In terms of evolution though, following this approach, the only time information that is available is the revision date.

Our analysis shows that most name evolution information is available in small passages within the current article of an entity. The corresponding excerpts in the texts have been identified by incorporating lists on Wikipedia, which provide semi-structured name evolution information for a limited set of entities. In future work, these findings will be used for extracting patterns and learning classifier models, which in turn can help to automatically discover more evolutions on Wikipedia as well as in other sources, like historical texts, newswire articles, social networks and blogs.

\section{Dataset}
\label{sec:dataset}

The data we used for our analysis was collected from the English Wikipedia on February 13, 2014. As starting point we used list pages dealing with name changes. Due to the different formats of these lists, we focused on those that are easy and reliable to parse. We found 19 lists in an appropriate format. Each name change on these is represented as follows: ``\textit{preceding name $\rightarrow$ succeeding name (date)}''). After filtering redundant items, we ended up with 10 lists. A downside of the format constraint is that we only found lists of geographic entities: \textit{Geographical renaming}, \textit{List of city name changes}, \textit{List of administrative division name changes} as well as lists dedicated to certain countries. The parsed lists contain 1,926 distinct entities with 2,852 name changes. For the found names, we fetched 2,782 articles. The larger number of articles compared to entities is a result of 766 entities with names that could be resolved to different articles. For 28 entities we were not able to resolve any name.

\section{Statistical Results}
\label{sec:statistics}

Based on the dataset we gathered statistics on name evolutions of entities in their corresponding articles. First, we analyzed the lists of entity name changes in terms of completeness and their suitability for such an analysis. Afterwards, we incorporated articles to analyze the mentions of the available changes.


\subsection{Lists of Entity Name Changes}

Out of the 1,926 entities with 2,852 name changes that we extracted from lists, 1,898 could be resolved to corresponding articles (98.5\%). As the ultimate goal is to identify excerpts that describe a full name change (i.e., consisting of preceding name, succeeding name and change date) only those that are annotated with corresponding change dates were subject of further analysis (e.g., ... $\rightarrow$ preceding $\rightarrow$ succeeding (date) $\rightarrow$ ...). Out of the remaining 2,810 changes, this holds for 918, which is 32.2\% of the ones that were originally extracted. These relevant changes belong to 696 entities, which is 36.1\% of all entities we started with. They constitute the subject of our research regarding excerpts that describe name evolutions.

\subsection{Mentions in Wikipedia Articles}

Proceeding with the entities that meet the prerequisites, we analyzed the corresponding articles for name changes mentioned in the text. For entities that were resolved to multiple articles, all articles have been taken into account. Out of the available 918 date-annotated name changes of entities with articles, we found 572 (62.3\%) mentioned with all three components (preceding name, succeeding name and change date). These were used for taking a closer look at the excerpts of texts that report changes. We measured the sentence distances of the three components for each of the 572 name changes. This is the minimum distance from a sentence mentioning one of the components to next sentence that mentions the last component, while bypassing the remaining one. For instance, if one sentence contains all the components, the sentence distance for this change is 0. Overall, the average sentence distance of the extracted excerpts was 19.9. However, this is caused by a very few, very high distances and is not representative, as indicated by the median of 1. In fact, 488 excerpts out of the total 572 excerpts, which is 85.3\%, have a distance less than 10. A significant majority of 79.7\% of these excerpts even have a sentence distance of less than three.

\section{Discussion}
\label{sec:discussion}

Our analysis was driven by the question about Wikipedia's suitability as a resource for extracting name evolution knowledge. The hypothesis was, that name evolutions are described in short excerpts within texts, which can be used later to learn common patterns to discover evolutions automatically. Based on our observations, this can be affirmed. More than 60\% of the 918 name changes that were available with corresponding articles and dates are mentioned in Wikipedia articles. Out of these, more than two-thirds were found within excerpts with less than three sentences. The extraction of particular patterns in order to train classifiers for the purpose of identifying evolutions automatically remains for future work. However, a first look at some excerpts already revealed that many of them contain certain signal words, such as ``became'', ``rename'', ``change''.

Unfortunately, the analyzed lists only consist of geographic entities as we were not able to fully reliably parse rather unstructured lists on Wikipedia that cover entities of different domains. At this point, accuracy was most important for building a foundation to train classifiers with a high precision in order to extend the training set later on. Accordingly, we can only carefully make the assumption that our observations hold for entities of other domains, too. This needs to be verified in future work. 

\section{Conclusions and Future Work}
\label{sec:conclusions}

In our study we investigated how name changes are mentioned in Wikipedia articles regardless of structural elements and found that a large majority is covered by short text passages. Using lists of name changes, we were able to automatically extract the corresponding excerpts from articles. Although the name evolutions mentioned in Wikipedia articles by far cannot be called complete, they provide a respectable basis for discovering more entity evolutions. In future work, we are going to use the excerpts that we found on Wikipedia for discovering patterns and training classifiers to find similar excerpts on further Wikipedia articles as well as other sources. The first step on this will be a more detailed analysis of the extracted excerpts, followed by engineering appropriate features. Eventually, we are going to build a knowledge base dedicated to entity evolutions. Such a knowledge base can serves as a source for application that rely on evolution knowledge, like information retrieval systems, especially on Web archives. Furthermore, it constitutes a ground truth for future research in the field of entity evolution, like novel algorithms for detecting entity evolutions on Web content streams.

\begin{flushleft}
\bibliographystyle{unsrtnat}
\bibliography{bib}{}

\begin{thebibliography}{4}
\providecommand{\natexlab}[1]{#1}
\providecommand{\url}[1]{\texttt{#1}}
\expandafter\ifx\csname urlstyle\endcsname\relax
  \providecommand{\doi}[1]{doi: #1}\else
  \providecommand{\doi}{doi: \begingroup \urlstyle{rm}\Url}\fi

\bibitem[Berberich et~al.(2009)Berberich, Bedathur, Sozio, and
  Weikum]{Berberich2009}
K.~Berberich, S.~J. Bedathur, M.~Sozio, and G.~Weikum.
\newblock Bridging the terminology gap in web archive search.
\newblock In \emph{WebDB}, 2009.

\bibitem[Kaluarachchi et~al.(2010)Kaluarachchi, Varde, Bedathur, Weikum, Peng,
  and Feldman]{Kaluarachchi2010}
A.~C. Kaluarachchi, A.~S. Varde, S.~J. Bedathur, G.~Weikum, J.~Peng, and
  A.~Feldman.
\newblock Incorporating terminology evolution for query translation in text
  retrieval with association rules.
\newblock In \emph{CIKM}, 2010.

\bibitem[Tahmasebi et~al.(2012)Tahmasebi, Gossen, Kanhabua, Holzmann, and
  Risse]{Tahmasebi2012}
N.~Tahmasebi, G.~Gossen, N.~Kanhabua, H.~Holzmann, and T.~Risse.
\newblock Neer: An unsupervised method for named entity evolution recognition.
\newblock In \emph{Coling}, Mumbai, India, 2012.
\newblock URL \url{http://www.l3s.de/neer-dataset}.

\bibitem[Kanhabua and Nørvåg(2010)]{Kanhabua2010}
Nattiya Kanhabua and Kjetil Nørvåg.
\newblock Exploiting time-based synonyms in searching document archives.
\newblock In \emph{JCDL}, 2010.

\end{thebibliography}
\end{flushleft}

\end{document}